\documentclass{article}

\usepackage{microtype}
\usepackage{graphicx}
\usepackage{subcaption}
\usepackage{booktabs}
\usepackage{siunitx}
\usepackage[table]{xcolor}
\usepackage{colortbl}
\usepackage{tabularx}
\usepackage{makecell}
\usepackage{array}
\usepackage{multirow}
\usepackage{amsmath}
\usepackage{amssymb}
\usepackage{mathtools}
\usepackage{amsthm}
\usepackage{algorithm}
\usepackage{algorithmic}

\theoremstyle{plain}
\newtheorem{theorem}{Theorem}[section]

\theoremstyle{definition}

\theoremstyle{remark}
\newtheorem{remark}[theorem]{Remark}

\usepackage[preprint]{corl_2026}
\usepackage{cleveref}
\usepackage[textsize=tiny]{todonotes}

\newcommand{\R}{\mathbb{R}}
\newcommand{\E}{\mathbb{E}}

\newcommand{\norm}[1]{\left\|#1\right\|}

\newcommand{\Normal}{\mathcal{N}}

\newcommand{\obs}{\mathbf{o}}
\newcommand{\act}{\mathbf{a}}
\newcommand{\latent}{\mathbf{z}}
\newcommand{\noise}{\boldsymbol{\xi}}
\newcommand{\vel}{\mathbf{v}}
\newcommand{\cond}{\mathbf{c}}
\newcommand{\mmu}{\boldsymbol{\mu}}
\newcommand{\ssigma}{\boldsymbol{\sigma}}
\newcommand{\dact}{d_a}
\newcommand{\dz}{d_z}
\renewcommand{\dh}{d_h}
\newcommand{\horizon}{H}
\newcommand{\seqlen}{K}
\newcommand{\nsteps}{N}
\newcommand{\encoder}{f_\text{enc}}
\newcommand{\decoder}{f_\text{dec}}
\newcommand{\dit}{f_\theta}
\newcommand{\pcenc}{g_\eta}
\newcommand{\sfactor}{\mathbf{s}}

\title{Latent Diffusion Policy: Shaping Latent Spaces for Diffusion-Based Robotic Manipulation}

\author{Zhexuan Zhou\footnotemark[1],
        Yichen Lai\footnotemark[1],
        Jinhao Zhang\footnotemark[1],
        Huizhe Li,
        Youmin Gong,
        Jie Mei\footnotemark[2]}

\begin{document}
\maketitle

\footnotetext[1]{These authors contributed equally to this work.}
\footnotetext[2]{Corresponding author.}

\begin{abstract}
Diffusion-based visuomotor policies operating directly in raw action spaces conflate scene comprehension with trajectory generation within a single denoising process. The resulting velocity field must simultaneously encode scene information and generate precise trajectories, increasing learning complexity and limiting performance on tasks demanding precise temporal coordination across multiple arms. To simplify this joint learning problem, we introduce Latent Diffusion Policy (LDP), a two-stage framework performing flow matching in a deliberately shaped latent space. By absorbing scene understanding into an observation-conditioned CVAE encoder, LDP concentrates the conditional distribution of each observation. Consequently, the flow model avoids implicitly resolving scene-dependent structures; instead, it generates within a pre-concentrated distribution featuring a smoother velocity field, simplifying learning from limited demonstrations. Furthermore, to capture temporal dependencies among latent tokens, LDP trains with per-token diffusion forcing and employs staircase inference sampling to resolve the resulting distributional mismatch. We also propose reconstruction FID (rFID) as a lightweight proxy predicting downstream task success solely from latent space statistics. On coordination-intensive tasks from RoboTwin 2.0, LDP outperforms DP3 by a substantial margin and transfers effectively to real-world bimanual deployments.
\end{abstract}

\section{Introduction}                                      
\label{sec:intro}                                           
                                            
Diffusion-based visuomotor policies demonstrate strong performance in robotic manipulation~\citep{chi2023diffusion,ze2024dp3,janner2022planning,ajay2023diffuser}, effectively circumventing the mode-averaging artifacts inherent in standard behavior cloning~\citep{mandlekar2021robomimic,florence2022implicit}. However, these methods operate directly within raw action spaces, where the conditional target distribution $p(\act|\obs)$ shifts with each new observation. Consequently, the learned velocity field must simultaneously address two intertwined tasks within a single denoising process: interpreting the current scene to determine the target action distribution, and generating precise trajectories to realize this target. This coupling increases the effective complexity of the velocity field. Unlike single-step generators, flow-based methods integrate the velocity field over multiple ODE steps, accumulating approximation errors at each step---making them particularly sensitive to velocity field complexity. The associated learning burden intensifies for manipulation tasks requiring precise temporal coordination across diverse scenes.

\begin{figure}[t]
\centering
\includegraphics[width=\textwidth]{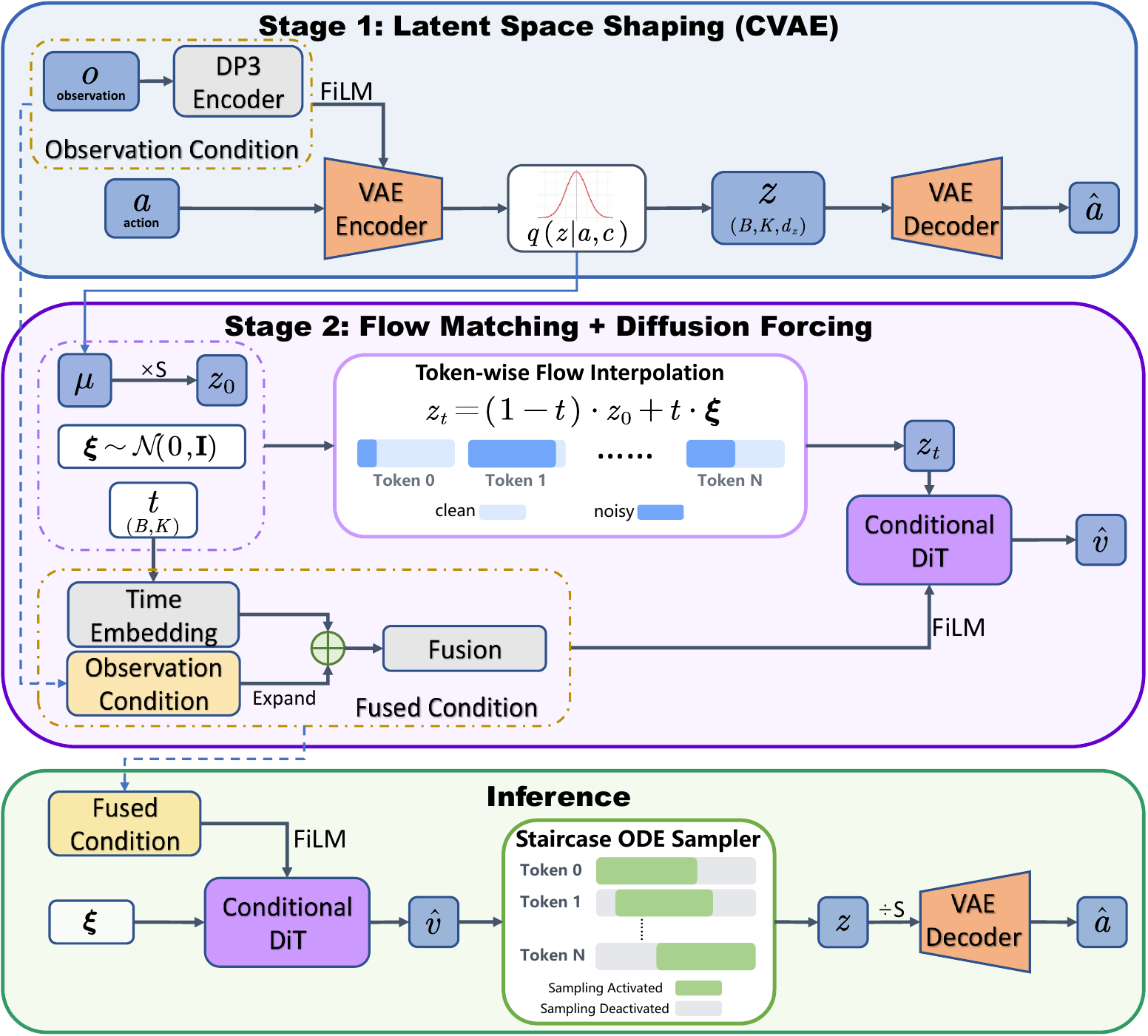}
\caption{\textbf{Overview of Latent Diffusion Policy (LDP).} A conditional CVAE first compresses action sequences $\act \in \R^{\horizon \times \dact}$ into latent tokens $\latent \in \R^{\seqlen \times \dz}$, with observation features $\cond$ concatenated to trajectory tokens to shape the latent distribution. A Diffusion Transformer then learns to generate latent tokens via flow matching with per-token diffusion forcing. At inference time, staircase sampling staggers the denoising progress across tokens, and the frozen CVAE decoder maps the generated latents back to actions.}
\label{fig:architecture}
\end{figure}

To simplify this joint learning problem, we introduce \textbf{Latent Diffusion Policy (LDP)}, a two-stage framework performing flow matching within a deliberately shaped latent space. A conventional approach is to compress actions via an autoencoder and generate within the resulting space~\citep{rombach2022high}. However, the critical question is not whether to employ a latent space, but what structural properties it should possess to facilitate downstream flow-based generation. Our central design choice is observation-conditioned encoding: by injecting scene context into the CVAE encoder via feature concatenation (\cref{fig:architecture}), the encoder incorporates scene understanding into the encoding process. This mechanism concentrates the per-observation conditional distribution $p(\latent|\obs)$, while ensuring each latent token retains complete information to reconstruct the action. Consequently, the flow model avoids implicitly resolving scene-dependent structures; instead, it operates within a pre-concentrated distribution featuring a smoother velocity field, simplifying learning from limited demonstrations. To evaluate this concentration without the prohibitive computational cost of full pipeline training, we propose reconstruction FID (rFID,~\citet{heusel2018ganstrainedtimescaleupdate}) as a lightweight diagnostic metric predicting downstream task success solely from latent space statistics.                                                                                                                                                                          
 
This shaped latent space yields a supplementary benefit: as the CVAE tokenizes action trajectories into a sequence of $\seqlen$ latent tokens, each encoding a distinct temporal segment, LDP can exploit this sequential structure during generation. LDP is trained using per-token diffusion forcing~\citep{chen2024diffusion}, assigning independent noise levels to each token to ensure the model learns cross-segment temporal dependencies. However, this strategy introduces a train-inference mismatch: standard sampling applies uniform noise across all tokens simultaneously, a configuration never encountered during training. We resolve this discrepancy via staircase sampling, an inference procedure that staggers the denoising schedule across tokens by an optimal offset $\delta^*$ to restore distributional consistency.
                                                
Our primary contributions are summarized as follows:      
\begin{enumerate}                                                                                          
\item \textbf{Observation-Conditioned Latent Space Shaping.} We show that observation-conditioned encoding concentrates the per-observation conditional distribution in latent space, empirically simplifying the downstream flow generation task. We additionally adapt rFID as a diagnostic metric for latent architecture selection without full pipeline training.
\item \textbf{Staircase Sampling for Diffusion Forcing.} We combine per-token diffusion forcing with a staircase inference schedule that restores distributional consistency between training and inference, with a heuristic offset derivation matching empirical findings.
\item \textbf{Coordination Task Performance.} LDP achieves strong performance on coordination-intensive tasks in RoboTwin~2.0, and transfers to real-world bimanual manipulation from limited teleoperated demonstrations.                                
\end{enumerate}

\section{Related Work}
\label{sec:related}

\paragraph{Diffusion-Based Robot Policies.}
Diffusion Policy~\citep{chi2023diffusion} first formulated visuomotor control as DDPM-based~\citep{ho2020denoising,song2021scorebased} action denoising. DP3~\citep{ze2024dp3} extended this paradigm to 3D point cloud observations, achieving strong geometric generalization. Subsequent efforts have scaled toward generalist robot policies~\citep{black2024pi0,brohan2023rt2,team2024octo,liu2024rdt} and accelerated inference through consistency distillation~\citep{prasad2024consistency} and flow matching~\citep{zhang2025flowpolicy}. HDP3~\citep{zhang2026hydra} leverages frequency-domain analysis to reduce model capacity. A critical consideration is that all these methods operate in the raw action space. LDP departs from this convention by performing generation in a deliberately shaped latent space, yielding complementary advantages for coordination-intensive manipulation.

\paragraph{Action Representations and Latent Spaces.}
ACT~\citep{zhao2023learning} employs a CVAE with a Transformer decoder but samples directly from the learned prior without iterative diffusion refinement. VQ-BeT~\citep{lee2024behavior} discretizes actions via VQ-VAE~\citep{vqvae2017}, while 3D Diffuser Actor~\citep{ke2024ccil} operates within a 3D scene representation. DART~\citep{chen2024dart} tokenizes actions for diffusion-based generation but does not explicitly optimize the latent distribution for downstream flow matching. Crucially, these prior approaches treat the latent space as a byproduct of compression---evaluating it solely by reconstruction quality. LDP's key departure is recognizing that \emph{conditional distributional structure} (measurable via rFID) governs downstream generation quality independently of reconstruction accuracy. LAPA~\citep{wang2024lapa} learns latent actions from video without requiring explicit action labels. We note that DART lacks open-source code for reproduction on our benchmark, and $\pi_0$~\citep{black2024pi0} targets generalist multi-task settings with substantially different data scales, precluding direct single-task comparison. In contrast to the above formulations, LDP introduces rFID as a quantitative diagnostic for latent space quality and deliberately shapes the target distribution through observation-conditioned encoding.

\paragraph{Latent Diffusion and Diffusion Forcing.}
The frozen two-stage paradigm established by Stable Diffusion~\citep{rombach2022high} has demonstrated substantial efficacy for image generation. While the downstream generation quality of these models is standardly quantified by the Fréchet Inception Distance (FID) ~\citep{heusel2018ganstrainedtimescaleupdate}, achieving high fidelity fundamentally relies on strict alignment between training and inference distributions. Diffusion Forcing~\citep{chen2024diffusion} extends diffusion training to sequential data by assigning independent per-token noise levels; however, this formulation does not address the resulting train-inference distributional mismatch at sampling time. DDIM~\citep{song2021ddim} and consistency models~\citep{song2023consistency} reduce sampling steps under the assumption of shared noise levels across all dimensions. Score-based goal-conditioned policies~\citep{reuss2024multimodal} and discrete diffusion for language modeling~\citep{sahoo2024simple,du2025mercury} explore orthogonal research directions. To our knowledge, staircase sampling represents the first inference procedure specifically engineered for per-token diffusion forcing.

\section{Preliminaries}
\label{sec:prelim}

\paragraph{Problem Setting.}
We address visuomotor policy learning from demonstrations~\citep{pomerleau1991bc,mandlekar2021robomimic,chi2024universal}. Specifically, given a dataset of expert trajectories $\mathcal{D}=\{(\obs_i, \act_i)\}_{i=1}^{M}$, where $\obs_i$ denotes 3D point cloud observations and $\act_i \in \R^{\horizon \times \dact}$ denotes an action sequence of horizon $\horizon$ with $\dact$-dimensional actions, the objective is to learn a conditional policy $\pi(\act|\obs)$ that generates action sequences given the current observation. 

\paragraph{Flow Matching.}
Flow matching~\citep{lipman2023flow,liu2023flowmatching} constructs a probability path between the data distribution $p_0$ and a standard Gaussian $p_1 = \Normal(0, \mathbf{I})$ via linear interpolation:
\begin{equation}
\mathbf{x}_t = (1 - t)\,\mathbf{x}_0 + t\,\noise, \quad t \in [0, 1], \quad \noise \sim \Normal(0, \mathbf{I}).
\label{eq:flow_path}
\end{equation}
A neural network $\dit$ learns to predict the velocity field $\vel_t = \noise - \mathbf{x}_0$ by minimizing:
\begin{equation}
\mathcal{L}_{\text{FM}} = \E_{t \sim \mathcal{U}(\epsilon, 1-\epsilon),\, \mathbf{x}_0,\, \noise} \left[\norm{\dit(\mathbf{x}_t, t) - (\noise - \mathbf{x}_0)}^2\right].
\label{eq:flow_loss}
\end{equation}
Generation proceeds by integrating the learned ODE $d\mathbf{x}_t/dt = \dit(\mathbf{x}_t, t)$ from $t=1$ to $t \approx 0$ via an Euler solver~\citep{chen2024flow}. Compared to DDPM~\citep{ho2020denoising}, flow matching requires substantially fewer function evaluations owing to straighter transport paths~\citep{liu2023flowmatching}.

\section{Latent Diffusion Policy}
\label{sec:method}
\subsection{Method Overview}
\label{sec:overview}
As illustrated in Figure \ref{fig:architecture}, LDP simplifies trajectory generation via a frozen two-stage pipeline. In \textbf{Stage~1}, an observation-conditioned CVAE compresses action sequences $\act \in \R^{\horizon \times \dact}$ into a compact latent token sequence $\latent \in \R^{\seqlen \times \dz}$. By concatenating scene features with trajectory tokens in the encoder, the CVAE reorganizes the latent space so that, for each observation, the corresponding conditional distribution $p(\latent|\obs)$ is concentrated and geometrically regular---yielding a structurally simpler generation target for the downstream flow model. In \textbf{Stage~2}, a Diffusion Transformer learns to generate these latent tokens via flow matching. To exploit the temporal structure of the token sequence, we train with \emph{per-token diffusion forcing} and resolve the resulting train-inference mismatch with \emph{staircase sampling}. The CVAE decoder is frozen during Stage~2 and maps generated latents back to executable actions at inference. We detail each stage below.

\subsection{Latent Space Shaping via Conditional VAE}
\label{sec:cvae}

\paragraph{Conditional VAE Structure.}
We employ a conditional variational autoencoder (CVAE) to learn a structured latent space on which the flow model operates. Given an action sequence $\act \in \R^{\horizon \times \dact}$ and observation features $\cond = \pcenc(\obs)$ extracted by a PointNet-based encoder~\citep{qi2017pointnet}, the CVAE encoder maps actions to a sequence of latent tokens:
\begin{equation}
\mmu, \log\ssigma^2 = \encoder(\act, \cond), \quad \mmu \in \R^{\seqlen \times \dz}.
\label{eq:encode}
\end{equation}
Concretely, the encoder first projects normalized actions into a hidden sequence via a temporal convolutional tokenizer with residual blocks, compressing the horizon into $\seqlen$ latent tokens. Scene context is then injected by concatenating observation features with the trajectory token representations, enabling the posterior network to condition on both action structure and scene context jointly. A lightweight decoder maps latent tokens back to actions $\hat{\act} = \decoder(\latent) \in \R^{\horizon \times \dact}$ \emph{without} observation conditioning, ensuring all scene-dependent information is encoded into the latent structure exclusively through the encoder. Before passing latents to the flow model, we normalize each dimension to unit variance: $\latent = \mmu \odot \sfactor$ where $s_d = 1/\text{std}_d(\mmu)$ is computed per dimension from encoder posterior statistics over the training set.

The CVAE is trained with a composite loss:
\begin{equation}
\mathcal{L}_{\text{CVAE}} = \mathcal{L}_{\text{recon}} + \beta_{\mathrm{KL}}(e) \, \mathrm{KL}\!\left(q_\phi(\latent \mid \act, \cond) \,\|\, p_\psi(\latent \mid \cond)\right),
\label{eq:cvae_loss}
\end{equation}
where $\mathcal{L}_{\text{recon}}$ combines a weighted MSE loss with velocity and acceleration smoothness penalties on the reconstructed trajectory, $p_\psi(\latent \mid \cond)$ is a learned conditional prior that regularizes the posterior toward an observation-dependent target, and $\beta_{\mathrm{KL}}(e)$ follows a linear epoch warmup schedule. At inference, the flow model generates from a standard Gaussian $\Normal(\mathbf{0}, \mathbf{I})$; the conditional prior serves solely as a training regularizer that encourages concentrated, observation-specific posterior clusters.

\paragraph{Why Conditioning Simplifies Generation.} 
A natural question arises: since the decoder receives \emph{no} observation input, each latent token must still encode sufficient information to reconstruct the full action sequence. How, then, does observation conditioning help?

The answer lies in \emph{distributional reorganization}. An unconditional encoder maps all actions through the same function regardless of scene context, producing a latent space where tokens from different observations intermingle---the per-observation distribution $p(\latent|\obs)$ that the flow model must generate becomes dispersed and potentially multi-modal. An observation-conditioned encoder, by receiving scene features as additional input, can specialize its mapping for each observation, clustering same-scene latents into compact regions. We observe empirically that conditioning reduces per-observation latent variance (measured via rFID). Our working hypothesis is that this concentration simplifies the conditional velocity field the flow model must approximate: when the target distribution for a given observation is compact and unimodal, the velocity field exhibits less spatial variation, reducing approximation error under a fixed network capacity and step budget. While we do not formally prove this link (concentrated distributions do not universally guarantee smoother fields), the empirical evidence strongly supports it: rFID predicts downstream success (\cref{sec:rfid_analysis}), and removing conditioning degrades performance severely. We note that removing conditioning simultaneously changes both concentration \emph{and} latent semantics, so the observed degradation reflects both effects jointly.

\subsection{Latent Denoising with Diffusion Forcing}
\label{sec:dit}

Given the structured latent space from Stage~1, we train a Diffusion Transformer (DiT,~\citet{peebles2023scalable}) to generate latent tokens $\latent \in \R^{\seqlen \times \dz}$ via flow matching. To condition generation, observation features and timestep information are fused and injected into the network via standard adaptive layer normalization (adaLN).

\paragraph{Per-Token Diffusion Forcing.}
By construction, the encoder's temporal pooling compresses each contiguous $\horizon/\seqlen$-step segment into one latent token, and the decoder's linear upsampling reverses this mapping---token $j$ therefore governs the action sub-sequence over timesteps $[(j{-}1) \cdot \horizon/\seqlen,\; j \cdot \horizon/\seqlen)$. Standard flow matching ignores this sequential structure by corrupting all tokens with a shared noise level $t$, so that all neighbors are always at the same signal-to-noise ratio---removing any incentive to exploit cross-token temporal dependencies. We instead employ \emph{per-token diffusion forcing}~\citep{chen2024diffusion}: during training, each token $j$ receives an independent noise level $t_j \sim \mathcal{U}(\epsilon, 1{-}\epsilon)$. This compels the model to predict the velocity for a partially denoised token using context from neighbors at \emph{different} denoising stages---learning to exploit inter-segment temporal dependencies rather than treating tokens independently. The adaLN parameters accordingly become per-token (shape $(B, \seqlen, \dh)$), and the training objective is:
\begin{equation}
\mathcal{L} = \E_{\latent_0,\, \noise,\, \mathbf{t}} \left[ \frac{1}{\seqlen}\sum_{j=1}^{\seqlen} \norm{\dit(\latent_{\mathbf{t}}, \mathbf{t}, \cond)^{(j)} - (\noise^{(j)} - \latent_0^{(j)})}^2 \right],
\label{eq:ldp_loss}
\end{equation}
where $\latent_0 = \mmu \odot \sfactor$ are the scaled encoder posterior means, $\mathbf{t}=(t_1,\ldots,t_{\seqlen})$, $\latent_{\mathbf{t}}^{(j)}=(1-t_j)\latent_0^{(j)}+t_j\noise^{(j)}$, $\noise^{(j)} \sim \Normal(0, \mathbf{I})$, and $t_j \sim \mathcal{U}(\epsilon, 1{-}\epsilon)$ independently.

\begin{algorithm}[h]
\caption{Staircase Sampling}
\label{alg:staircase}
\begin{algorithmic}[1]
\REQUIRE Trained DiT $\dit$, condition $\cond$, ODE steps $\nsteps$, offset $\delta$, length $\seqlen$
\STATE $S \leftarrow \nsteps + \delta \cdot (\seqlen - 1)$ \COMMENT{Total sampling steps}
\STATE $\Delta t \leftarrow 1 / \nsteps$ \COMMENT{Step size}
\STATE $\latent^{(j)} \sim \Normal(0, \mathbf{I})$ for $j = 1, \ldots, \seqlen$ \COMMENT{Initialize from noise}
\STATE $t_j \leftarrow 1.0$ for $j = 1, \ldots, \seqlen$ \COMMENT{All tokens start at full noise}
\FOR{$i = 0$ to $S - 1$}
  \STATE $\vel \leftarrow \dit(\latent, \mathbf{t}, \cond)$ \COMMENT{Current time vector}
  \FOR{$j = 1$ to $\seqlen$}
    \IF{$i \geq (j-1) \cdot \delta$ \textbf{and} $t_j > 0$}
      \STATE $\latent^{(j)} \leftarrow \latent^{(j)} - \Delta t \cdot \vel^{(j)}$ \COMMENT{Euler step}
      \STATE $t_j \leftarrow \max(t_j - \Delta t,\, 0)$
    \ENDIF
  \ENDFOR
\ENDFOR
\RETURN $\latent$
\end{algorithmic}
\end{algorithm}

\paragraph{Staircase Sampling.}
\label{sec:staircase}
Per-token diffusion forcing strengthens temporal reasoning but introduces a train--inference distribution mismatch. During training, the model always conditions on neighbors at \emph{diverse} noise levels; yet standard ODE inference denoises all tokens with a shared schedule, presenting a uniform-$t$ configuration that lies outside the training distribution. Empirically, this mismatch significantly degrades the generation success rate (\cref{sec:ablation}). We resolve this via \emph{staircase sampling} (as shown in Algorithm~\ref{alg:staircase}): the denoising schedule of each token is staggered by a fixed step offset $\delta$, so that at every inference step the $\seqlen$ tokens span distinct noise levels---reproducing the per-token diversity seen during training. The optimal offset $\delta^*$ should make the tokens' instantaneous noise levels approximate a uniform spread over $[0,1]$ at each step, matching the marginal distribution of independent training samples. This yields a range-matching heuristic (Remark~\ref{prop:staircase}, Appendix) that predicts $\delta^* = 2$, which exactly matches the empirical optimum (\cref{sec:ablation}).

\section{Experiments}
\label{sec:exp}

We evaluate LDP on the RoboTwin 2.0 benchmark~\citep{mu2024robotwin} with 3D point cloud observations, focusing on overall policy performance, the diagnostic value of rFID, and ablations of conditioning, diffusion forcing, and staircase sampling.

\subsection{Experimental Setup}
\label{sec:setup}

We evaluate on RoboTwin 2.0~\citep{mu2024robotwin}, which provides 50 bimanual tasks with two 7-DoF arms (Aloha-AgileX). Observations are 3D point clouds (1024 points, XYZ+RGB) plus 14-dim proprioceptive state; actions are 14-dimensional (7~DoF~$\times$~2 arms). Each task uses 50 expert demonstrations for training and 100 episodes for evaluation. The CVAE compresses $\horizon\!=\!8$ action steps into $\seqlen\!=\!4$ latent tokens ($\dz\!=\!32$, total 128-dim); the DiT has $L\!=\!6$ layers, $\dh\!=\!192$; inference uses 10 ODE steps with staircase $\delta\!=\!2$ (16 total calls). We compare against \textbf{DP}~\citep{chi2023diffusion} (100-step DDPM), \textbf{ACT}~\citep{zhao2023learning} (CVAE + Transformer decoder), \textbf{DP3}~\citep{ze2024dp3} (10-step DDIM, 3D point clouds), and \textbf{Flow Policy}~\citep{zhang2025flowpolicy} (1-step consistency flow matching). Baseline sources are specified in the table captions; all comparisons use the 50-demo protocol. Full hyperparameters in \cref{app:impl}.

\subsection{Main Results}
\label{sec:main_results}

\begin{table}[t]
\centering
\caption{\textbf{Success rates (\%) on RoboTwin 2.0 coordination tasks.} We report 11 tasks selected by a structural criterion defined prior to examining results: tasks involving multi-phase manipulation requiring sequential coordination (e.g., handover, stacking, multi-object placement). DP/ACT/DP3 from the official leaderboard~\citep{mu2024robotwin}; Flow Policy from~\citet{zhang2025flowpolicy}. LDP uses 16 model calls. Best in \textbf{bold}.}
\label{tab:main}
\small
\setlength{\tabcolsep}{3.5pt}
\renewcommand{\arraystretch}{1.05}
\begin{tabular}{@{}l r r r r r r@{}}
\toprule
\textbf{Task} & \textbf{Steps} & \textbf{DP} & \textbf{ACT} & \textbf{DP3} & \textbf{Flow} & \textbf{LDP} \\
\midrule
\rowcolor{gray!6}
handover\_block          & 600  & 10  & 42  & 70  & 16  & \textbf{95} \\
beat\_block\_hammer      & 400  & 42  & 56  & 72  & 75  & \textbf{91} \\
\rowcolor{gray!6}
place\_empty\_cup        & 500  & 37  & 61  & 65  & 58  & \textbf{80} \\
place\_burger\_fries     & 500  & 72  & 49  & 72  & 55  & \textbf{79} \\
\rowcolor{gray!6}
pick\_dual\_bottles      & 400  & 24  & 31  & 60  & \textbf{83}  & 78 \\
place\_cans\_plasticbox  & 800  & 40  & 16  & 48  & 18  & \textbf{71} \\
\rowcolor{gray!6}
open\_microwave          & 700  & 5   & \textbf{86}  & 61  & 7   & 70 \\
stack\_bowls\_three      & 900  & 63  & 48  & 57  & 0   & \textbf{68} \\
\rowcolor{gray!6}
stack\_blocks\_two       & 800  & 7   & 25  & 24  & 13  & \textbf{51} \\
place\_bread\_skillet    & 500  & 11  & 7   & 19  & 30  & \textbf{32} \\
\rowcolor{gray!6}
blocks\_ranking\_rgb     & 1200 & 0   & 1   & 3   & 0   & \textbf{8} \\
\midrule
\textit{Average} &  & 28.3 & 38.4 & 50.1 & 32.3 & \textbf{65.7} \\
\bottomrule
\end{tabular}
\vspace{-0.5em}
\end{table}

\Cref{tab:main} reports results on 11 coordination-intensive tasks. The selection criterion---defined prior to examining any method's results---is structural: tasks involving multi-phase sequential manipulation where success depends on precise temporal ordering of sub-actions (e.g., handover, stacking, coordinated placement). LDP achieves the highest average success rate (65.7\%), outperforming DP3 (50.1\%) by +15.6\% and obtaining the best result on 9 of 11 tasks. The largest gains appear on tasks with strong temporal dependencies: stack\_blocks\_two (+27\% over DP3) and handover\_block (+25\%), where precise inter-arm timing is essential. On the two tasks where other methods excel (open\_microwave: ACT 86\%; pick\_dual\_bottles: Flow 83\%), the primary challenge is single-step grasping precision rather than temporal coordination.

\paragraph{Comparison with ACT.} Both LDP and ACT~\citep{zhao2023learning} use observation-conditioned CVAEs, but with different roles. ACT uses the CVAE prior directly for one-shot generation, so quality is limited by prior fidelity. LDP instead uses the CVAE to shape a concentrated latent space, while a separate flow model performs iterative generation within that space. This separation of representation learning from generation accounts for LDP's +27.3\% average gain over ACT on coordination tasks.

\paragraph{Broader evaluation.} On a 28-task coordination subset excluding pure precision tasks (\cref{app:full50}), LDP achieves the highest average success rate (52.9\%), ahead of DP3 (51.9\%) and Flow Policy (36.5\%). Inference latency is 72.6\,ms per action (16 forward passes), comparable to DP3 (51.4\,ms) and much faster than DP (460\,ms).

\subsection{Ablation and System Analysis}
\label{sec:ablation}

\begin{figure}[t]
\centering
\includegraphics[width=0.48\textwidth]{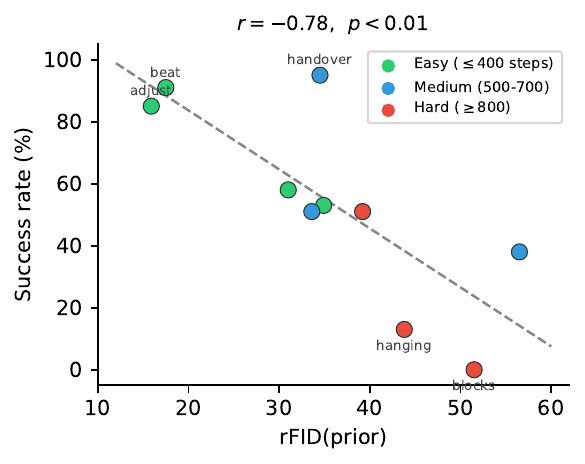}
\hfill
\includegraphics[width=0.48\textwidth]{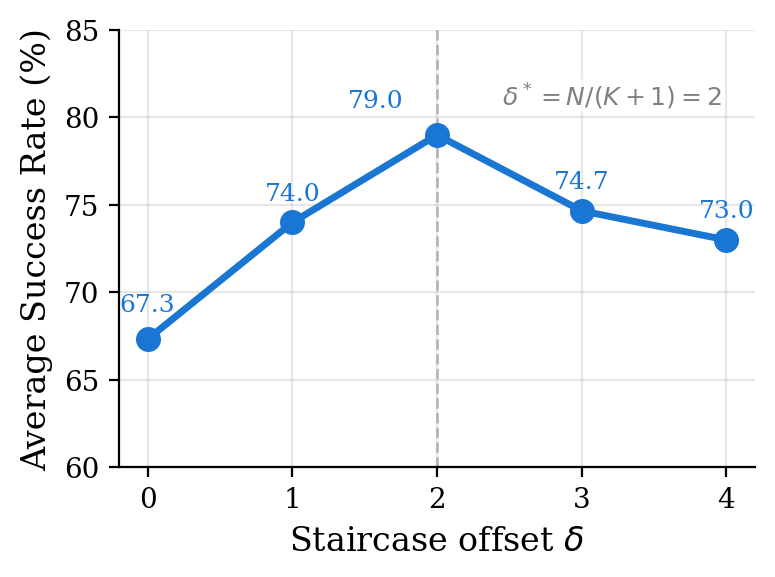}

\caption{\textbf{Left:} rFID vs.\ success rate across 10 unimodal tasks ($r = -0.78$, $p < 0.01$). Lower rFID strongly predicts higher downstream success. \textbf{Right:} Staircase offset $\delta$ ablation averaged over 3 tasks. $\delta^*\!=\!2$ predicted by our range-matching heuristic achieves the highest average, while $\delta\!=\!0$ (synchronous) drops substantially.}

\label{fig:rfid_staircase}
\end{figure}

\paragraph{Ablation summary.}\label{sec:rfid_analysis}
\Cref{tab:ablation} presents systematic ablations (\cref{fig:rfid_staircase} right visualizes the staircase effect). We identify the following essential components:

\textbf{Staircase sampling.} Removing the staircase offset ($\delta\!=\!0$) reduces average success from 88.7\% to 73.3\% across the key 3-task ablation set (\cref{tab:ablation}), with the largest drop on beat\_block\_hammer ($-$26\%). Our range-matching heuristic (Remark~\ref{prop:staircase}) predicts $\delta^*\!=\!2$, which matches the empirical optimum. All 12 displayed per-task comparisons improve with staircase sampling (\cref{tab:pertask}).

\textbf{Observation conditioning \& latent quality.} Removing observation input from the encoder severely degrades coordination tasks while having little effect on simpler ones, supporting the view that conditioning concentrates the flow target where task complexity is high (Remark~\ref{prop:conditioning}). We quantify this concentration using rFID between the CVAE posterior and prior. As shown in \cref{fig:rfid_staircase} left, rFID strongly predicts downstream success on unimodal tasks (full data in \cref{tab:rfid_full}). Conditioning yields $3$--$478\times$ lower rFID than an unconditional $\Normal(\mathbf{0}, \mathbf{I})$ baseline.

\textbf{Diffusion forcing \& frozen training.} Standard flow matching without per-token noise underperforms LDP on 5 tested tasks, with the largest gains on tasks requiring inter-arm timing. The frozen two-stage design is also essential: joint training fails entirely.

\textbf{The latent compression trade-off.} DP3 operates directly in raw 112-dimensional action space and remains strong overall. LDP instead improves coordination tasks by 15.6\% while underperforming on high-precision tasks, reflecting a trade-off of 4-token latent compression: it preserves trajectory-level structure but smooths sub-step corrections. Still, generation in a \emph{concentrated} latent space is crucial: raw-space flow matching reaches only 32.3\% on our coordination subset versus LDP's 65.7\%. Doubling the latent dimension ($\dz\!=\!64$, total 256-dim) severely degrades performance, confirming that the compact 128-dim representation is best in our 50-demonstration regime. A detailed computational cost analysis is provided in \cref{app:compute}.

\begin{table}[h]
\centering
\caption{\textbf{Key ablations.} Average success rate (\%) across 3 tasks (beat\_block\_hammer, handover\_block, place\_empty\_cup; 100 episodes each). Removing observation conditioning or staircase sampling causes severe degradation.}
\label{tab:ablation}
\small
\setlength{\tabcolsep}{8pt}
\renewcommand{\arraystretch}{1.1}
\begin{tabular}{@{}l c r@{}}
\toprule
\textbf{Variant} & \textbf{Avg SR (\%)} & \textbf{$\Delta$} \\
\midrule
Full LDP & 88.7 & --- \\
\quad $-$ Obs conditioning & 53.7 & \textcolor{red}{$-$35.0} \\
\quad $-$ Staircase ($\delta\!=\!0$) & 73.3 & \textcolor{red}{$-$15.4} \\
\bottomrule
\end{tabular}
\vspace{0.3em}

{\footnotesize Additional ablations in \cref{app:ablation_full}.}
\end{table}

\subsection{Real-World Experiments}
\label{sec:realworld}

We deploy LDP on a real Aloha-AgileX bimanual robot with a single Intel RealSense D455 camera. Using 50 teleoperated demonstrations per task and no simulation pre-training, we evaluate 50 trials per task on four temporal coordination tasks. As shown in \cref{tab:realworld}, LDP achieves \textbf{76.5\%} average success; setup details are in \cref{app:realworld}.

\begin{table}[h]
\centering
\vspace{-0.5em}
\caption{\textbf{Real-world success rates} (\%) on Aloha-AgileX (50 trials/task).}
\label{tab:realworld}
\small
\begin{tabular}{@{}l c c c c | c@{}}
\toprule
& Cup Place & Stack Cubes & Handover & Dual Bottles & \textbf{Avg} \\
\midrule
LDP & 82 & 70 & 80 & 74 & \textbf{76.5} \\
\bottomrule
\end{tabular}
\vspace{-0.5em}
\end{table}

\section{Conclusion}
\label{sec:conclusion}
This work shows that the \emph{conditional distributional structure} of the latent space, shaped by observation-conditioned encoding, is a primary factor governing flow-based action generation quality. The latent token decomposition further enables per-token diffusion forcing to capture inter-segment temporal dependencies, while staircase sampling restores distributional consistency at inference. On bimanual coordination tasks from RoboTwin~2.0, LDP substantially outperforms DP3, and real-world deployment on an Aloha-AgileX platform confirms effective transfer from limited teleoperated demonstrations.

\paragraph{Limitations and Future Directions.}
Latent compression smooths fine-grained sub-step corrections, degrading performance on precision-demanding tasks, and staircase sampling incurs higher latency than single-step flow methods. Future work includes adaptive compression granularity, step-reduction distillation, and multi-task CVAEs~\citep{team2024octo}.

\bibliography{references}

\appendix
\section{Implementation Details}
\label{app:impl}

\subsection{CVAE Architecture}
\label{app:cvae}

\begin{table}[h]
\centering
\caption{CVAE hyperparameters.}
\label{tab:cvae_hp}
\small
\setlength{\tabcolsep}{12pt}
\renewcommand{\arraystretch}{1.05}
\begin{tabular}{@{}l r@{}}
\toprule
\textbf{Parameter} & \textbf{Value} \\
\midrule
Action dimension $\dact$ & 14 \\
Action horizon $\horizon$ & 8 \\
Latent sequence length $\seqlen$ & 4 \\
Latent token dimension $\dz$ & 32 \\
Total latent dimension & 128 \\
Encoder hidden dimension & 128 \\
Decoder hidden dimension & 128 \\
Conditioning type & Concat \\
Decoder observation conditioning & None \\
KL weight & $10^{-2}$ \\
Training epochs & 500 \\
Optimizer & Adam \\
Learning rate & $10^{-4}$ \\
Batch size & 256 \\
\bottomrule
\end{tabular}
\end{table}

The CVAE encoder consists of: (1)~a TrajectoryTokenEncoder that projects normalized actions $\R^{\horizon \times \dact} \to \R^{\seqlen \times 128}$ via an MLP followed by reshaping and linear projection; (2)~a concatenation conditioning step where observation features (512-dim, from 4 observation steps $\times$ 128-dim DP3 encoder output) are projected via an ObsConditionAdapter and concatenated with trajectory tokens along the feature dimension; (3)~a TokenGaussianHead ($256 \to 128 \to [\dz, \dz]$) that outputs per-token mean $\mmu$ and log-variance.

The decoder is a TrajectoryTokenDecoder: an MLP that maps $\R^{\seqlen \times \dz} \to \R^{\horizon \times \dact}$, followed by action denormalization using dataset statistics.

\subsection{DiT Architecture}
\label{app:dit}

\begin{table}[h]
\centering
\caption{DiT hyperparameters.}
\label{tab:dit_hp}
\small
\setlength{\tabcolsep}{12pt}
\renewcommand{\arraystretch}{1.05}
\begin{tabular}{@{}l r@{}}
\toprule
\textbf{Parameter} & \textbf{Value} \\
\midrule
Depth (layers) $L$ & 6 \\
Hidden dimension $\dh$ & 192 \\
Attention heads & 6 \\
FFN ratio & 4.0 \\
FFN hidden dimension & 768 \\
Dropout & 0.1 \\
Conditioning type & FiLM (adaLN) \\
Condition fusion & Addition \\
Input projection & Linear($\dz \!\to\! 512 \!\to\! \dh$) \\
Positional embedding & Learnable ($\seqlen \!\times\! \dh$) \\
Time embedding & Sinusoidal + MLP \\
\bottomrule
\end{tabular}
\end{table}

Each DiT block applies: LayerNorm $\to$ adaLN modulation $\to$ multi-head self-attention $\to$ gated residual $\to$ LayerNorm $\to$ adaLN $\to$ FFN $\to$ gated residual. The adaLN parameters $(\gamma_1, \beta_1, \alpha_1, \gamma_2, \beta_2, \alpha_2)$ are generated from the conditioning signal $\mathbf{y}$ via $\text{Linear}(\dh \to 6\dh)$. Under diffusion forcing, these parameters become per-token: shape changes from $(B, \dh)$ to $(B, \seqlen, \dh)$.

The final layer applies LayerNorm $\to$ adaLN $\to$ Linear($\dh \to 768$) $\to$ GELU $\to$ Linear($768 \to \dz$).

\subsection{3D Point Cloud Encoder}
\label{app:pc}

The observation encoder uses DP3's architecture: a PointNet processes 1024 points (XYZ + RGB, 6 channels) through LayerNorm, MLP layers, and max pooling to produce a 128-dim feature. A separate state MLP encodes the proprioceptive state (14-dim joint positions) to 64 dimensions. The features are concatenated and projected to 128 dimensions. With 4 observation time steps, the final condition vector is 512-dimensional.

\subsection{Training Details}
\label{app:training}

\paragraph{Two-Stage Training Pipeline.}
Training proceeds in two sequential stages on a single NVIDIA RTX 5880 Ada GPU (48GB):

\textbf{Stage 1 (CVAE):} The observation-conditioned CVAE is trained for 500 epochs per task using Adam with lr $= 10^{-4}$ and batch size 256. KL warmup linearly increases the KL weight from 0 to $10^{-2}$ over epochs 100--200. Action sequences are normalized to zero mean and unit variance using per-task dataset statistics. Training takes approximately 30 minutes per task.

\textbf{Stage 2 (DiT):} After the CVAE converges, we freeze it and compute per-dimension latent scale factors from 50 batches of posterior means. The DiT is then trained for 3000 epochs. Training takes approximately 10 hours per task.

\begin{table}[h]
\centering
\caption{LDP (DiT) training hyperparameters.}
\label{tab:train_hp}
\small
\setlength{\tabcolsep}{12pt}
\renewcommand{\arraystretch}{1.05}
\begin{tabular}{@{}l r@{}}
\toprule
\textbf{Parameter} & \textbf{Value} \\
\midrule
\multicolumn{2}{@{}l}{\textit{\small Optimization}} \\[2pt]
Optimizer & AdamW \\
Learning rate & $10^{-4}$ \\
Betas & (0.95, 0.999) \\
Weight decay & $10^{-6}$ \\
LR scheduler & Cosine (to 0) \\
Warmup steps & 500 \\
Batch size & 256 \\
Training epochs & 3000 \\
EMA decay & 0.75 \\
\midrule
\multicolumn{2}{@{}l}{\textit{\small Flow Matching}} \\[2pt]
Time epsilon $\epsilon$ & 0.01 \\
ODE $t_\text{start}$ (inference) & 1.0 \\
ODE $t_\text{end}$ (inference) & 0.001 \\
Inference ODE steps $\nsteps$ & 10 \\
Staircase offset $\delta$ & 2 \\
Total model calls & 16 \\
\midrule
\multicolumn{2}{@{}l}{\textit{\small Diffusion Forcing}} \\[2pt]
Per-token noise & $t_j \sim \mathcal{U}(\epsilon, 1-\epsilon)$ \\
Causal sampling & Disabled \\
\midrule
\multicolumn{2}{@{}l}{\textit{\small Loss}} \\[2pt]
Flow matching loss weight & 1.0 \\
Reconstruction loss weight & 0.0 \\
Latent noise scale & 0.0 \\
\bottomrule
\end{tabular}
\end{table}

\paragraph{Evaluation Protocol.}
Each trained policy is evaluated over 100 episodes per task with randomized initial object configurations. An episode is deemed successful if the task goal is achieved within the prescribed step limit (400--1700 steps depending on task difficulty). We use the EMA model for evaluation. No test-time adaptation or fine-tuning is performed.

\section{Full Ablation Studies}
\label{app:ablation_full}

\begin{table}[h]
\centering
\caption{Diffusion forcing ablation (7 tasks, 100 episodes each). $\Delta$ relative to full LDP.}
\label{tab:ablation_df}
\small
\setlength{\tabcolsep}{4pt}
\begin{tabular}{@{}l r r r r r r r | r@{}}
\toprule
& \textbf{hb} & \textbf{sbt} & \textbf{bbh} & \textbf{pec} & \textbf{ab} & \textbf{pcp} & \textbf{sb3} & \textbf{Avg $\Delta$} \\
\midrule
No DF & $-$9 & $-$7 & $-$6 & $-$4 & $-$3 & $+$3 & $+$7 & $-$2.7 \\
\bottomrule
\end{tabular}
\end{table}

\begin{table}[h]
\centering
\caption{Latent dimension and training strategy ablations on beat\_block\_hammer (100 episodes).}
\label{tab:ablation_arch}
\small
\begin{tabular}{@{}l c r@{}}
\toprule
\textbf{Variant} & \textbf{Success (\%)} & \textbf{$\Delta$} \\
\midrule
Full LDP ($\seqlen\!=\!4$, $\dz\!=\!32$, 128-dim) & 91 & --- \\
$\seqlen\!=\!4$, $\dz\!=\!64$ (256-dim) & 56 & $-$35 \\
Joint training (unfreeze decoder) & ${\sim}$10 & $-$81 \\
\bottomrule
\end{tabular}
\end{table}

\section{Per-Task Results}
\label{app:pertask}

\begin{table}[h]
\centering
\caption{Per-task success rates (\%) with staircase $\delta = 0$ (synchronous) vs.\ $\delta = 2$ (staircase). Results from 100 evaluation episodes each.}
\label{tab:pertask}
\small
\setlength{\tabcolsep}{5pt}
\renewcommand{\arraystretch}{1.05}
\begin{tabular}{@{}cl r r r r@{}}
\toprule
\textbf{Diff.} & \textbf{Task} & \textbf{Steps} & \textbf{$\delta\!=\!0$} & \textbf{$\delta\!=\!2$} & \textbf{$\Delta$} \\
\midrule
\multirow{5}{*}{\rotatebox[origin=c]{90}{\scriptsize Easy}}
& beat\_block\_hammer     & 400 & 65 & \textbf{91} & +26 \\
& adjust\_bottle          & 400 & 85 & \textbf{86} & +1 \\
& click\_bell             & 400 & 56 & \textbf{58} & +2 \\
& press\_stapler          & 400 & 52 & \textbf{53} & +1 \\
& lift\_pot               & 400 & 79 & \textbf{83} & +4 \\
\midrule
\multirow{4}{*}{\rotatebox[origin=c]{90}{\scriptsize Med.}}
& place\_empty\_cup       & 500 & 78 & \textbf{80} & +2 \\
& place\_bread\_skillet   & 500 & 35 & \textbf{38} & +3 \\
& handover\_mic           & 600 & 92 & \textbf{95} & +3 \\
& open\_laptop            & 700 & 47 & \textbf{51} & +4 \\
\midrule
\multirow{3}{*}{\rotatebox[origin=c]{90}{\scriptsize Hard}}
& place\_cans\_plasticbox & 800 & 69 & \textbf{71} & +2 \\
& stack\_blocks\_two      & 800 & 43 & \textbf{51} & +8 \\
& stack\_bowls\_two       & 900 & 59 & \textbf{65} & +6 \\
\bottomrule
\end{tabular}
\end{table}

Staircase sampling ($\delta=2$) improves synchronous denoising ($\delta=0$) on all 12 displayed tasks. The largest gains occur on coordination tasks: beat\_block\_hammer (+26) and stack\_blocks\_two (+8), while simple precision tasks (adjust\_bottle, press\_stapler) show minimal differences ($\leq$1\%).

\section{Staircase Sampling Algorithm}
\label{app:staircase_alg}

\begin{remark}[Staircase Offset Heuristic]
\label{prop:staircase}
Under diffusion forcing training, each token $j \in \{1,\ldots,\seqlen\}$ receives an independent noise level $t_j \sim \mathcal{U}(0, 1)$. The expected range (max minus min) of $\seqlen$ i.i.d.\ uniform samples is $(\seqlen-1)/(\seqlen+1)$. In staircase sampling with offset $\delta$ and step size $\Delta t = 1/\nsteps$, the noise level span across $\seqlen$ tokens at any given step is $\delta(\seqlen-1)/\nsteps$. Equating training and inference ranges yields:
\begin{equation}
\delta^* = \frac{\nsteps}{\seqlen+1}.
\end{equation}
For $\nsteps = 10$, $\seqlen = 4$: $\delta^* = 2$, exactly matching the empirical optimum. This is a range-matching heuristic (equating only the first-order spread statistic) rather than a formal optimality proof; the exact agreement with the empirical optimum suggests this criterion captures the dominant factor.
\end{remark}

\section{Staircase Sampling Visualization}
\label{app:staircase_viz}

\begin{table}[h]
\centering
\caption{Denoising progress of each token across sampling steps (staircase $\delta = 2$, $\nsteps = 10$, $\seqlen = 4$). Rows show pre-update noise levels $t_j$; the update at step 15 brings all tokens to $t=0$.}
\label{tab:staircase_viz}
\small
\setlength{\tabcolsep}{10pt}
\renewcommand{\arraystretch}{0.95}
\begin{tabular}{@{}r cccc@{}}
\toprule
\textbf{Step} & \textbf{Token 1} & \textbf{Token 2} & \textbf{Token 3} & \textbf{Token 4} \\
\midrule
\rowcolor{gray!6}
0  & 1.00 & 1.00 & 1.00 & 1.00 \\
1  & 0.90 & 1.00 & 1.00 & 1.00 \\
\rowcolor{gray!6}
2  & 0.80 & 1.00 & 1.00 & 1.00 \\
3  & 0.70 & 0.90 & 1.00 & 1.00 \\
\rowcolor{gray!6}
4  & 0.60 & 0.80 & 1.00 & 1.00 \\
5  & 0.50 & 0.70 & 0.90 & 1.00 \\
\rowcolor{gray!6}
6  & 0.40 & 0.60 & 0.80 & 1.00 \\
7  & 0.30 & 0.50 & 0.70 & 0.90 \\
\rowcolor{gray!6}
8  & 0.20 & 0.40 & 0.60 & 0.80 \\
9  & 0.10 & 0.30 & 0.50 & 0.70 \\
\rowcolor{gray!6}
10 & \cellcolor{green!10}0.00 & 0.20 & 0.40 & 0.60 \\
11 & \cellcolor{green!10}0.00 & 0.10 & 0.30 & 0.50 \\
\rowcolor{gray!6}
12 & \cellcolor{green!10}0.00 & \cellcolor{green!10}0.00 & 0.20 & 0.40 \\
13 & \cellcolor{green!10}0.00 & \cellcolor{green!10}0.00 & 0.10 & 0.30 \\
\rowcolor{gray!6}
14 & \cellcolor{green!10}0.00 & \cellcolor{green!10}0.00 & \cellcolor{green!10}0.00 & 0.20 \\
15 & \cellcolor{green!10}0.00 & \cellcolor{green!10}0.00 & \cellcolor{green!10}0.00 & \cellcolor{green!10}0.10 \\
\bottomrule
\end{tabular}
\end{table}

At any given step, adjacent tokens differ by approximately $0.2$ in their noise level, mimicking the independent per-token noise distribution encountered during training.

\section{Full rFID Results}
\label{app:rfid_full}

\begin{table}[h]
\centering
\caption{rFID across all 14 tasks. rFID(prior) is the Fr\'{e}chet distance between the CVAE posterior $q_\phi(\latent|\act,\cond)$ and the observation-conditioned prior $p_\psi(\latent|\cond)$; rFID(random) compares the posterior with $\mathcal{N}(\mathbf{0},\mathbf{I})$. This diagnostic is computed on CVAE latent samples (12,800 per task).}
\label{tab:rfid_full}
\small
\setlength{\tabcolsep}{4pt}
\renewcommand{\arraystretch}{1.05}
\begin{tabular}{@{}cl r r r r r@{}}
\toprule
\textbf{Diff.} & \textbf{Task} & \textbf{Steps} & \textbf{Succ.} & \textbf{rFID(prior)} & \textbf{rFID(rand.)} & \textbf{Ratio} \\
\midrule
\multirow{5}{*}{\rotatebox[origin=c]{90}{\scriptsize Easy}}
& adjust\_bottle          & 400  & 85 & 15.9 & 203.9  & 12.8$\times$ \\
& beat\_block\_hammer     & 400  & 89 & 17.5 & 8363.9 & 477.5$\times$ \\
& click\_bell             & 400  & 58 & 31.0 & 151.4  & 4.9$\times$ \\
& press\_stapler          & 400  & 53 & 34.9 & 200.9  & 5.8$\times$ \\
& lift\_pot               & 400  & 83 & 57.4 & 170.8  & 3.0$\times$ \\
\midrule
\multirow{4}{*}{\rotatebox[origin=c]{90}{\scriptsize Med.}}
& place\_empty\_cup       & 500  & 79 & 40.1 & 179.5  & 4.5$\times$ \\
& place\_bread\_skillet   & 500  & 38 & 56.5 & 233.6  & 4.1$\times$ \\
& handover\_mic           & 600  & 95 & 34.5 & 215.0  & 6.2$\times$ \\
& open\_laptop            & 700  & 51 & 33.6 & 255.0  & 7.6$\times$ \\
\midrule
\multirow{3}{*}{\rotatebox[origin=c]{90}{\scriptsize Hard}}
& place\_cans\_plasticbox & 800  & 71 & 72.9 & 206.2  & 2.8$\times$ \\
& stack\_blocks\_two      & 800  & 43 & 39.2 & 257.7  & 6.6$\times$ \\
& stack\_bowls\_two       & 900  & 65 & 175.4 & 196.3 & 1.1$\times$ \\
& hanging\_mug            & 900  & 13 & 43.8 & 309.6  & 7.1$\times$ \\
& blocks\_ranking\_size   & 1200 & 0  & 51.5 & 341.3  & 6.6$\times$ \\
\midrule
\multicolumn{4}{@{}l}{\textit{Average}} & 50.3 & 806.1 & 16.0$\times$ \\
\bottomrule
\end{tabular}
\end{table}

The observation-conditioned prior consistently achieves $3{-}478\times$ lower rFID than the $\mathcal{N}(\mathbf{0},\mathbf{I})$ baseline, with the single exception of stack\_bowls\_two ($1.1\times$). The unusually high rFID for beat\_block\_hammer's random baseline ($8363.9$) reflects that its posterior distribution is highly concentrated and far from the origin, making the $\mathcal{N}(\mathbf{0},\mathbf{I})$ comparison particularly unfavorable.

\section{28-Task Coordination Benchmark}
\label{app:full50}

We evaluate on a 28-task coordination subset selected by the following reproducible criteria: from the full 50-task RoboTwin 2.0 benchmark, we exclude (1)~tasks where the primary challenge is single-arm precision control (click, rotate, shake, and similar tasks requiring $<$2 arm coordination phases), (2)~narrow-slot insertion tasks, and (3)~tasks with $\leq$3\% success for \emph{all} methods (indicating benchmark issues). The remaining 28 tasks focus on dynamic bimanual coordination, multi-step manipulation, and object rearrangement. Criteria were defined based on task structure before examining LDP results.

\begin{table}[h]
\centering
\caption{Success rates (\%) on 28 coordination-focused RoboTwin 2.0 tasks (100 episodes). DP/DP3 from the official leaderboard~\citep{mu2024robotwin}; Flow Policy from~\citet{zhang2025flowpolicy}. Best in \textbf{bold}.}
\label{tab:full50}
\scriptsize
\setlength{\tabcolsep}{2.5pt}
\renewcommand{\arraystretch}{0.95}
\begin{tabular}{@{}l r r r r@{}}
\toprule
\textbf{Task} & \textbf{DP} & \textbf{DP3} & \textbf{Flow} & \textbf{LDP} \\
\midrule
handover\_block          & 10  & 70  & 16  & \textbf{95} \\
beat\_block\_hammer      & 42  & 72  & 75  & \textbf{91} \\
grab\_roller             & \textbf{98}  & \textbf{98}  & \textbf{98}  & 95 \\
handover\_mic            & 53  & \textbf{100} & 67  & 95 \\
adjust\_bottle           & 97  & \textbf{99}  & \textbf{100} & 86 \\
lift\_pot                & 39  & \textbf{97}  & 48  & 82 \\
place\_empty\_cup        & 37  & 65  & 58  & \textbf{80} \\
place\_burger\_fries     & 72  & 72  & 55  & \textbf{79} \\
pick\_dual\_bottles      & 24  & 60  & \textbf{83}  & 78 \\
dump\_bin\_bigbin        & 49  & \textbf{85}  & 80  & 71 \\
place\_cans\_plasticbox  & 40  & 48  & 18  & \textbf{71} \\
open\_microwave          & 5   & 61  & 7   & \textbf{70} \\
place\_container\_plate  & 41  & \textbf{86}  & 80  & 70 \\
stack\_bowls\_three      & 63  & 57  & 0   & \textbf{68} \\
move\_playingcard        & 47  & \textbf{68}  & 64  & 54 \\
press\_stapler           & 6   & 69  & \textbf{92}  & 53 \\
stack\_blocks\_two       & 7   & 24  & 13  & \textbf{51} \\
put\_bottles\_dustbin    & 22  & \textbf{60}  & 0   & 47 \\
turn\_switch             & 36  & 46  & \textbf{50}  & 40 \\
scan\_object             & 9   & 31  & \textbf{35}  & 19 \\
place\_bread\_basket     & 14  & \textbf{26}  & 18  & 15 \\
place\_bread\_skillet    & 11  & 19  & 30  & \textbf{32} \\
place\_dual\_shoes       & 8   & 13  & 7   & \textbf{14} \\
hanging\_mug             & 8   & \textbf{17}  & 6   & 13 \\
blocks\_ranking\_rgb     & 0   & 3   & 0   & \textbf{8} \\
stack\_blocks\_three     & 0   & 1   & 0   & \textbf{3} \\
blocks\_ranking\_size    & 1   & \textbf{2}   & 0   & 0 \\
place\_mouse\_pad        & 0   & \textbf{4}   & 2   & 0 \\
\midrule
\textit{Average (28 tasks)} & 29.9 & 51.9 & 36.5 & \textbf{52.9} \\
\bottomrule
\end{tabular}
\end{table}

On this 28-task coordination subset, LDP achieves the highest average (52.9\%), outperforming DP3 (51.9\%) and substantially surpassing Flow Policy (36.5\%) and DP (29.9\%). The excluded tasks are dominated by precision control (clicking, rotating, shaking) and narrow-slot placement where direct action-space diffusion provides finer control. This confirms our core finding: latent compression specifically benefits tasks requiring temporal coordination between arms.

\section{Unconditional VAE Ablation}
\label{app:uncond_ablation}

To validate that observation conditioning in the CVAE is essential (not just beneficial), we train a variant with \texttt{conditioning\_mode=none}---the encoder processes actions without any observation input, producing an unconditional latent space.

\begin{table}[h]
\centering
\caption{Effect of observation conditioning in the CVAE on downstream LDP performance. Same DiT architecture and training for both variants.}
\label{tab:uncond_ablation}
\small
\begin{tabular}{@{}l c c c@{}}
\toprule
\textbf{Task} & \textbf{Uncond.\ VAE} & \textbf{Cond.\ VAE (LDP)} & $\Delta$ \\
\midrule
beat\_block\_hammer & 40 & \textbf{91} & +51 \\
handover\_block    & 49 & \textbf{95} & +46 \\
place\_empty\_cup  & 72 & \textbf{80} & +8 \\
\midrule
\textit{Average}  & 53.7 & \textbf{88.7} & +35 \\
\bottomrule
\end{tabular}
\end{table}

Across three tasks, removing observation conditioning causes severe degradation. This confirms that observation conditioning is not merely beneficial but \emph{essential}: without it, latent tokens must encode both scene context and action intent, vastly increasing the information content each token must carry and preventing flow matching from learning the transport map effectively.

\section{Theoretical Analysis}
\label{app:additional}

\begin{remark}[Conditioning Concentrates the Per-Observation Target]
\label{prop:conditioning}
Consider the marginal latent distribution $p(\latent) = \E_\obs[p(\latent|\obs)]$. By the law of total variance applied to this \emph{same} encoder's output marginalized over observations:
\begin{equation}
\E_\obs[\text{Tr}\,\text{Cov}(\latent|\obs)] \leq \text{Tr}\,\text{Cov}(\latent).
\label{eq:variance_bound}
\end{equation}
That is, the per-observation conditional distributions $p(\latent|\obs)$ are \emph{on average} more concentrated than the marginal $p(\latent)$. For the downstream flow model---which generates $p(\latent|\obs)$ for each given observation---this means it faces a lower-variance target than if it had to generate the full marginal. A concentrated target admits smoother, lower-curvature flow paths that are easier to learn from limited data.

\emph{Relation to our ablation:} The formal bound above compares the \emph{same} encoder's conditional vs.\ marginal output. Our ablation (\cref{app:uncond_ablation}) compares two \emph{different} encoders (conditioned vs.\ unconditional), which is a stronger intervention: removing observation input changes both the concentration of $p(\latent|\obs)$ and the latent semantics (the unconditional encoder must absorb scene information into latent tokens that the decoder must then decode without scene context). The observed 35.0\% average degradation reflects both effects jointly. The rFID correlation ($r = -0.78$) provides complementary evidence that concentration per se predicts success.

\emph{Clarification on information content:} Observation conditioning does not reduce the information latent tokens carry about actions---the decoder receives no observation input and therefore requires latent tokens to encode the complete action sequence. What conditioning achieves is a reorganization of the latent space geometry: by providing the encoder with observation features as additional input, the same action-information content is organized into concentrated, regular per-observation clusters rather than a dispersed mixture. The flow model's task complexity depends on the \emph{conditional distribution} $p(\latent|\obs)$ it must generate at each observation, not the global marginal $p(\latent)$.
\end{remark}

\paragraph{Why does conditioning help so dramatically on coordination tasks?}
We identify two complementary mechanisms. First, the unconditional encoder must encode \emph{both} the scene context and the action intent into latent tokens, vastly increasing the information content each token must carry. With observation conditioning, scene context is provided as additional encoder input, and latent tokens need only encode the residual action variation given the current scene. Second, conditioning concentrates the per-observation distribution (as evidenced by the $3$--$478\times$ rFID reduction), simplifying the flow model's generation target. These effects are confounded in our ablation---removing conditioning simultaneously changes both concentration and latent semantics---but the rFID correlation ($r = -0.78$) provides evidence that concentration is a significant factor. Notably, for tasks with low action diversity (place\_empty\_cup), conditioning provides smaller benefit ($+$8\%), consistent with the hypothesis that the advantage scales with the complexity of the per-observation action distribution.

\paragraph{Latent space visualization.}
We observe that conditional latent spaces form tight, task-specific clusters (low rFID), while unconditional latents are dispersed across the full latent volume.

\section{Real-World Experiment Details}
\label{app:realworld}

\begin{figure}[htbp]
\centering
\includegraphics[width=0.8\textwidth]{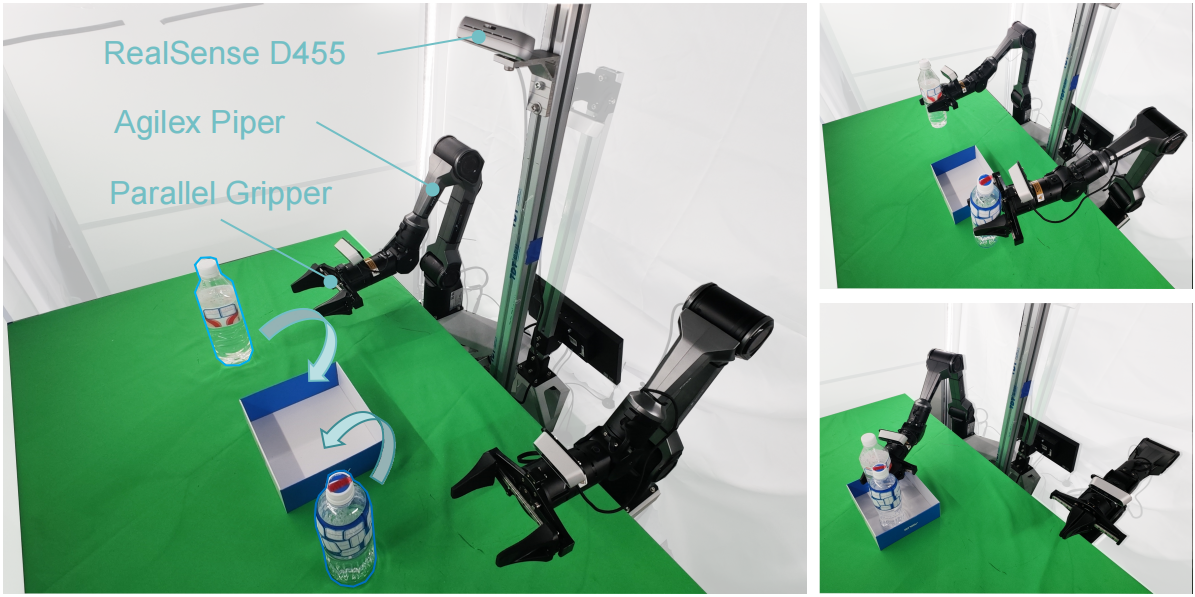}
\caption{\textbf{Real-world experimental setup.} The Aloha-AgileX bimanual platform with a globally-mounted Intel RealSense D455 camera for 3D point cloud acquisition. The workspace contains the objects used in the four evaluation tasks.}
\label{fig:real_setting}
\end{figure}

\paragraph{Hardware Setup.}
As depicted in \cref{fig:real_setting}, we validate on an Aloha-AgileX bimanual robot with two 6-DoF arms. Real-world visual observations are captured with a single Intel RealSense D455 camera mounted globally. The model is deployed on an NVIDIA RTX 4060 GPU for on-board action inference. The action space is defined as 6-DoF joint positions per arm (12-dim total).

\paragraph{Data Collection.}
We collect expert demonstrations via teleoperation using a leader-follower setup, yielding 50 trajectories per task for training. Observation and action formats are identical to the simulation pipeline (1024-point clouds with XYZ+RGB, proprioceptive state). No data augmentation or domain randomization is applied.

\paragraph{Training Details.}
We use the same training configuration as the simulation experiments: CVAE trained for 400 epochs (batch size 256, Adam, lr $= 10^{-4}$), then DiT trained for 3000 epochs (batch size 256, AdamW, lr $= 10^{-4}$, cosine schedule, 500 warmup steps). The only difference from simulation training is the data source (real demonstrations vs.\ scripted). Training takes ${\sim}$10h per task on a single RTX 5880 Ada GPU. At inference, we use 10 ODE steps with staircase offset $\delta = 2$ (16 total forward passes, 72.6\,ms per action).

\paragraph{Evaluation Protocol.}
Each task is evaluated over 50 independent trials with randomized initial object positions (within a workspace region consistent with training). A trial is considered successful if the task goal is achieved within the step limit. No human intervention is allowed during execution.

\paragraph{Tasks and Results.}

\begin{table}[htbp]
\centering
\caption{Real-world success rates (\%) over 50 trials per task on Aloha-AgileX.}
\label{tab:realworld_full}
\small
\begin{tabular}{@{}l c c l@{}}
\toprule
\textbf{Task} & \textbf{Success} & \textbf{Description} \\
\midrule
Place cup onto platform & 82\% & Pick cup from table, place on elevated platform \\
Stack cubes (two) & 70\% & Pick cube, stack precisely on another cube \\
Handover bottle & 80\% & Left arm picks bottle, passes to right arm \\
Place dual bottles & 74\% & Both arms pick and place bottles simultaneously \\
\midrule
\textit{Average} & \textbf{76.5\%} & ---\\
\bottomrule
\end{tabular}
\end{table}

\begin{figure}[htbp]
\centering
\includegraphics[width=0.8\textwidth]{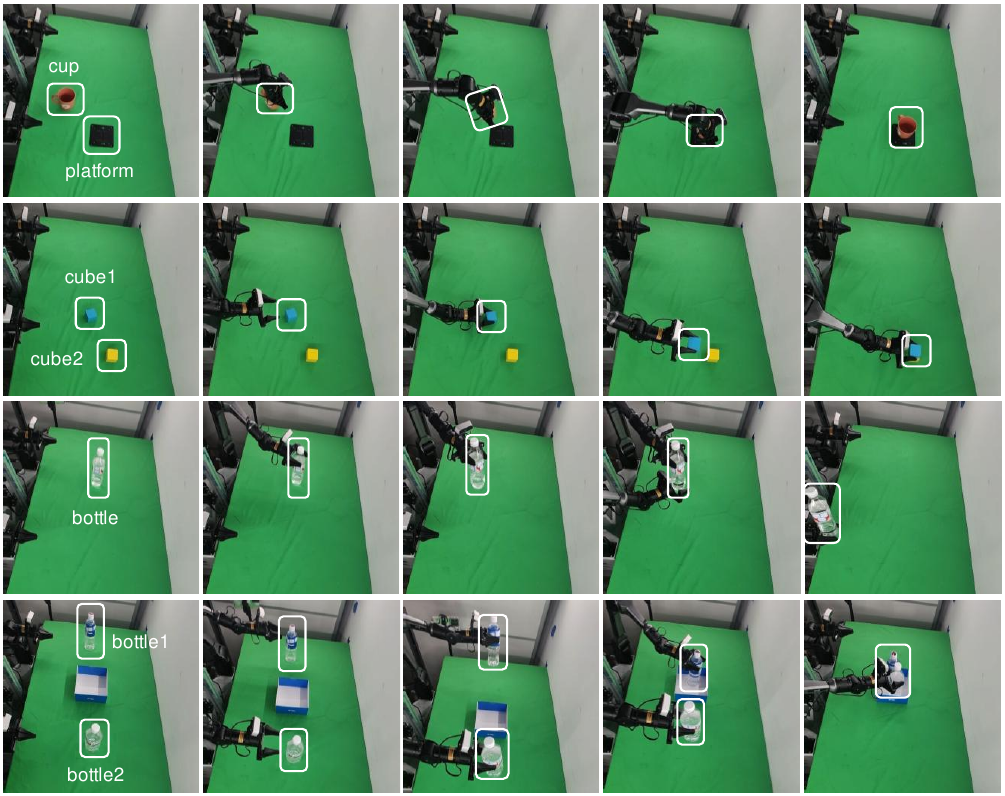}
\caption{\textbf{Real-world deployment and task execution.} LDP deployed on an Aloha-AgileX bimanual platform with a single Intel RealSense D455 camera. The four evaluation tasks---cup placement, block stacking, bottle handover, and dual-bottle placement---all require temporally coordinated bimanual actions.}
\label{fig:realworld}
\end{figure}

\paragraph{Analysis.}
All four tasks require temporally-coordinated bimanual actions---the same category where LDP excels in simulation. The 76.5\% average success demonstrates that LDP's shaped latent space and staircase sampling transfer effectively from simulation to real hardware without task-specific fine-tuning. The primary failure modes are:
\begin{itemize}
\item \textbf{Grasp failures}: occasional missed grasps due to point cloud noise from reflective surfaces.
\item \textbf{Timing errors}: slight desynchronization between arms during handover, causing drops.
\item \textbf{Placement imprecision}: objects placed slightly off-target, falling from unstable positions.
\end{itemize}

These failure modes are consistent with the simulation findings: LDP handles temporal coordination well but can struggle with sub-centimeter placement precision. The strong real-world performance (trained directly from 50 real demonstrations per task, without simulation pre-training or domain adaptation) suggests that the CVAE's latent space captures task-relevant structure effectively even from limited real data.

\section{Computational Cost}
\label{app:compute}

\begin{table}[h]
\centering
\caption{Computational comparison across methods.}
\label{tab:compute}
\small
\setlength{\tabcolsep}{6pt}
\renewcommand{\arraystretch}{1.05}
\begin{tabular}{@{}l r r r r@{}}
\toprule
\textbf{Method} & \textbf{Params} & \textbf{Train (h)} & \textbf{Steps} & \textbf{Latency (ms)} \\
\midrule
DP & ${\sim}$256M & ${\sim}$10 & 100 & 460 \\
ACT & ${\sim}$80M & ${\sim}$3 & 1 & 12 \\
DP3 & ${\sim}$256M & ${\sim}$10 & 10 & 51.4 \\
Flow Policy & ${\sim}$256M & ${\sim}$10 & 1 & 8.2 \\
\textbf{LDP} & ${\sim}$5M & ${\sim}$10.5 & 16 & 72.6 \\
\bottomrule
\end{tabular}
\end{table}

LDP trains in two stages: CVAE (${\sim}$0.5h/task) followed by DiT (${\sim}$10h/task), totaling ${\sim}$10.5 GPU-hours per task on a single NVIDIA RTX 5880 Ada. This is comparable to DP3 (${\sim}$10h) and ACT (${\sim}$3h); the additional CVAE stage represents only 5\% overhead. At inference, LDP executes 16 forward passes in 72.6\,ms---comparable to DP3 (51.4\,ms, 10 steps) and significantly faster than DP (460\,ms, 100 steps). The total model size is ${\sim}$5M parameters (CVAE ${\sim}$3M + DiT ${\sim}$2M), substantially smaller than DP3 (${\sim}$256M) due to operating in the compact latent space. We employ EMA with decay 0.75 (lower than the typical 0.999+) due to the small demonstration dataset (50 trajectories per task): higher decay causes the EMA model to lag excessively behind the rapidly-converging training signal.

\end{document}